\documentclass{svproc}

\usepackage{amsmath}
\usepackage{amssymb}
\usepackage{graphicx}
\usepackage{url}

\usepackage{siunitx}
\usepackage{array}
\newcolumntype{P}[1]{>{\centering\arraybackslash}p{#1}}
\usepackage{subcaption}

\begin{document}

\mainmatter
\title{Estimation of Payload Inertial Parameters from Human Demonstrations by Hand Guiding}
\titlerunning{Payload Inertial Parameters from Human Demonstrations}
\author{Johannes Hartwig \and
Philipp Lienhardt \and
Dominik Henrich}
\authorrunning{J. Hartwig et al.} 
\institute{
Chair for Applied Computer Science III (Robotics and Embedded Systems)\\ 
University of Bayreuth, D-95440 Bayreuth, Germany\\
\email{johannes.hartwig@uni-bayreuth.de}}
\maketitle
\begin{abstract}

As the availability of cobots increases, it is essential to address the needs of users with little to no programming knowledge to operate such systems efficiently. 
Programming concepts often use intuitive interaction modalities, such as hand guiding, to address this. 
When programming in-contact motions, such frameworks require knowledge of the robot tool's payload inertial parameters (PIP) in addition to the demonstrated velocities and forces to ensure effective hybrid motion-force control. 
This paper aims to enable non-expert users to program in-contact motions more efficiently by eliminating the need for a dedicated PIP calibration, thereby enabling flexible robot tool changes. 
Since demonstrated tasks generally also contain motions with non-contact, our approach uses these parts to estimate the robot's PIP using established estimation techniques. 
The results show that the estimation of the payload's mass is accurate, whereas the center of mass and the inertia tensor are affected by noise and a lack of excitation. 
Overall, these findings show the feasibility of PIP estimation during hand guiding but also highlight the need for sufficient payload accelerations for an accurate estimation.

\end{abstract}

\section{Introduction} 
\label{sec:introduction}

The increasing availability of cobots has highlighted the need for task experts, especially those without programming knowledge, to be able to program and reconfigure robot systems easily \cite{Perzylo2019,Riedelbauch2019}. 
Programming by demonstration (PbD), particularly through hand guiding, offers a promising approach by allowing users to teach robot trajectories without prior programming experience of robots \cite{Villani2018}. 
With this approach, users can directly guide a robot to complete a task using only their domain knowledge. 
Enhanced playback programming systems, for example, enable such intuitive interactions \cite{Riedl2021}.
Various tasks in automation, handling, and robotics require a precise specification not only of position over time but also of force over time, so-called in-contact motions \cite{Suomalainen2022,Hartwig2023b}.
Measuring the forces between the robot's end effector and the environment is essential to program such motions effectively.
A widely used solution, also utilized in this paper, incorporates a force/torque sensor (FTS) between the user's hand-guiding point and the robot's tool \cite{Montebelli2015,Steinmetz2015,Conkey2019}. 
This setup enables the simultaneous recording of the robot's trajectory and the forces exerted on the environment during hand-guided demonstrations.

Handling various tasks requires diverse tools gripped by the end effector in small and medium-sized enterprises (SMEs).
However, variable payload data emerging from these different tools introduces challenges in the programming and execution phases. 
For example, inconsistent payload inertial parameters (PIP) can complicate segmentation into different motion phases (e.g., guarded and compliant motions) as it is unclear if these forces arise because of interaction with the environment or inertia. 
This can cause problems during the programming phase and disturb control strategies during task execution. 
Therefore, we need information about the robot's PIP, namely mass, center of mass, and inertia tensor, to eliminate their contribution to the measured forces and torques \cite{Hollerbach2016}.  
While this is not an issue if computer-aided design (CAD) and material modeling are available, such models are not always present, and modeling can be complex for non-experts.
In such cases, methods to determine these parameters automatically through optimization are often employed.
These inertial parameters optimization techniques typically utilize special calibration motions that ensure the necessary acceleration of the payload and other conditions, such as minimizing the influence of noise in the sensor data \cite{Kubus2008}.
However, in the context of PbD, it would be beneficial to determine these parameters with less specific, ideally even arbitrary movements. 
This would allow for an estimation during the user's demonstration and a sped-up programming process.

Hence, this paper examines two key issues:
(i) First, it explores to what extent payload parameters can be estimated using non-specialized trajectories derived from human demonstrations.
(ii) Second, it investigates which optimization technique provides useful results within a reasonable time frame.
For this, we compare three PIP optimization methods using three motions with both synthetic validation data and real-world measured data.

This paper is structured as follows:
Section \ref{sec:related_work} presents some existing approaches for determining the physical parameters of objects. 
Section \ref{sec:methodology} describes the method implemented and the corresponding experimental setup. 
Section \ref{sec:results} discusses the results of the experiments, especially with respect to the deviation of the calculated load data from the ground truth parameters modeled by hand. 
Finally, Section \ref{sec:conclusion} concludes the paper and discusses possible future research and the associated challenges.

\section{Optimization Problem and Related Work} 
\label{sec:related_work}

The determination of the inertial parameters, including mass $m \in \mathbb{R}$, the center of mass $c \in \mathbb{R}^3$, and the symmetric inertia tensor $I \in \mathbb{R}^{3 \times 3}$, is typically achieved by measuring force $f \in \mathbb{R}^3$ and torque $\tau \in \mathbb{R}^3$. 
These measurements are complemented by the observed linear velocities $v \in \mathbb{R}^3$ and angular velocities $\omega \in \mathbb{R}^3$, as well as the respective accelerations $a \in \mathbb{R}^3$ and $\alpha \in \mathbb{R}^3$.
The Newton-Euler equations of motion describe the relationship between all these variables \cite{Kubus2008,Kozlowski1998}:
\begin{equation}
\begin{split}
    f &= m \cdot a + m \cdot g + \alpha \times m \cdot c + \omega \times (\omega \times m \cdot c) \\
    \tau &= I \cdot \alpha + \omega \times (I \cdot \omega) + m \cdot c \times a + m \cdot c \times g
\label{eq:newton}
\end{split}
\end{equation}

\begin{figure}[tb]
    \centering
    \includegraphics[width=0.49\linewidth]{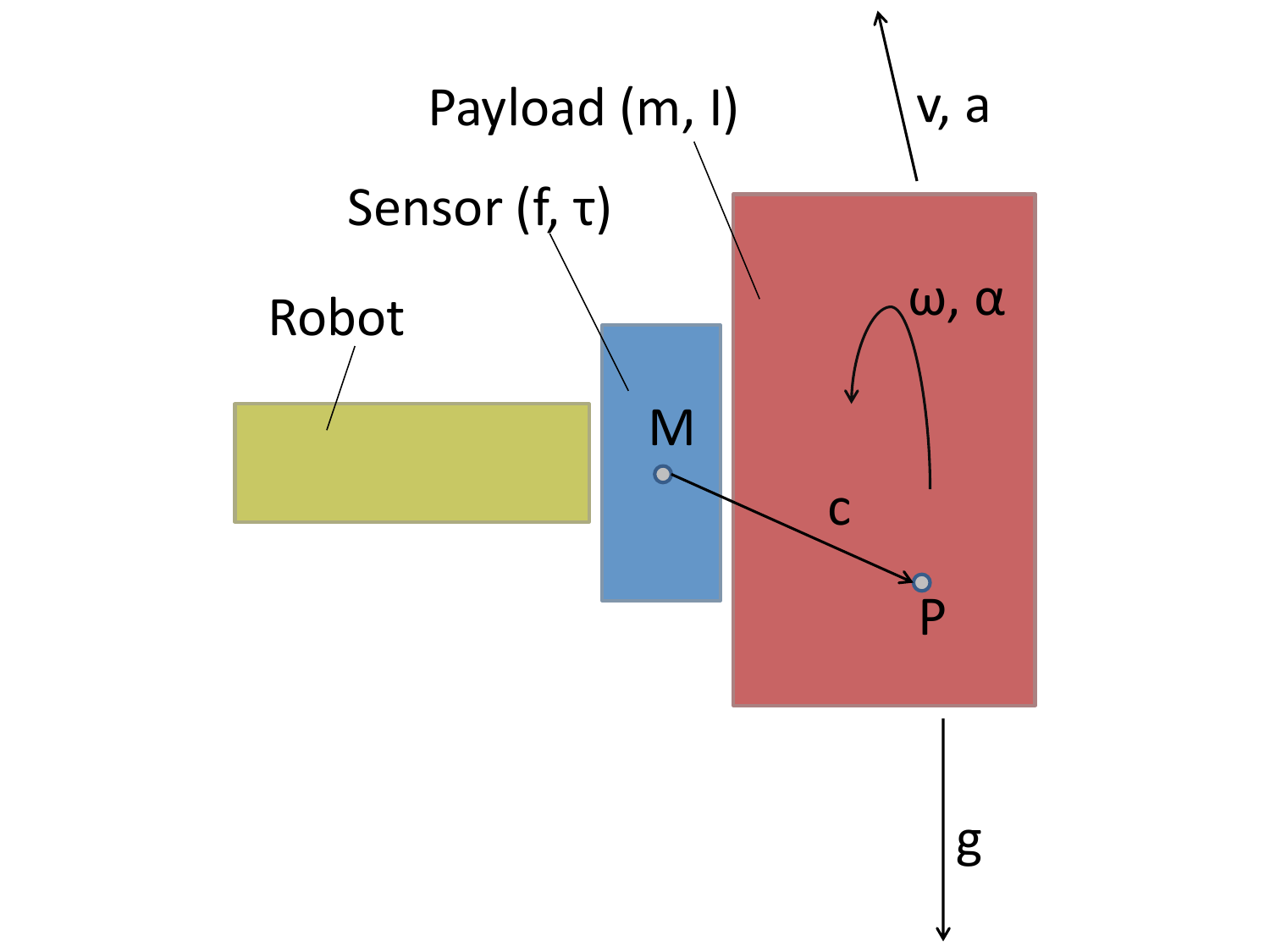}
    \includegraphics[width=0.49\linewidth]{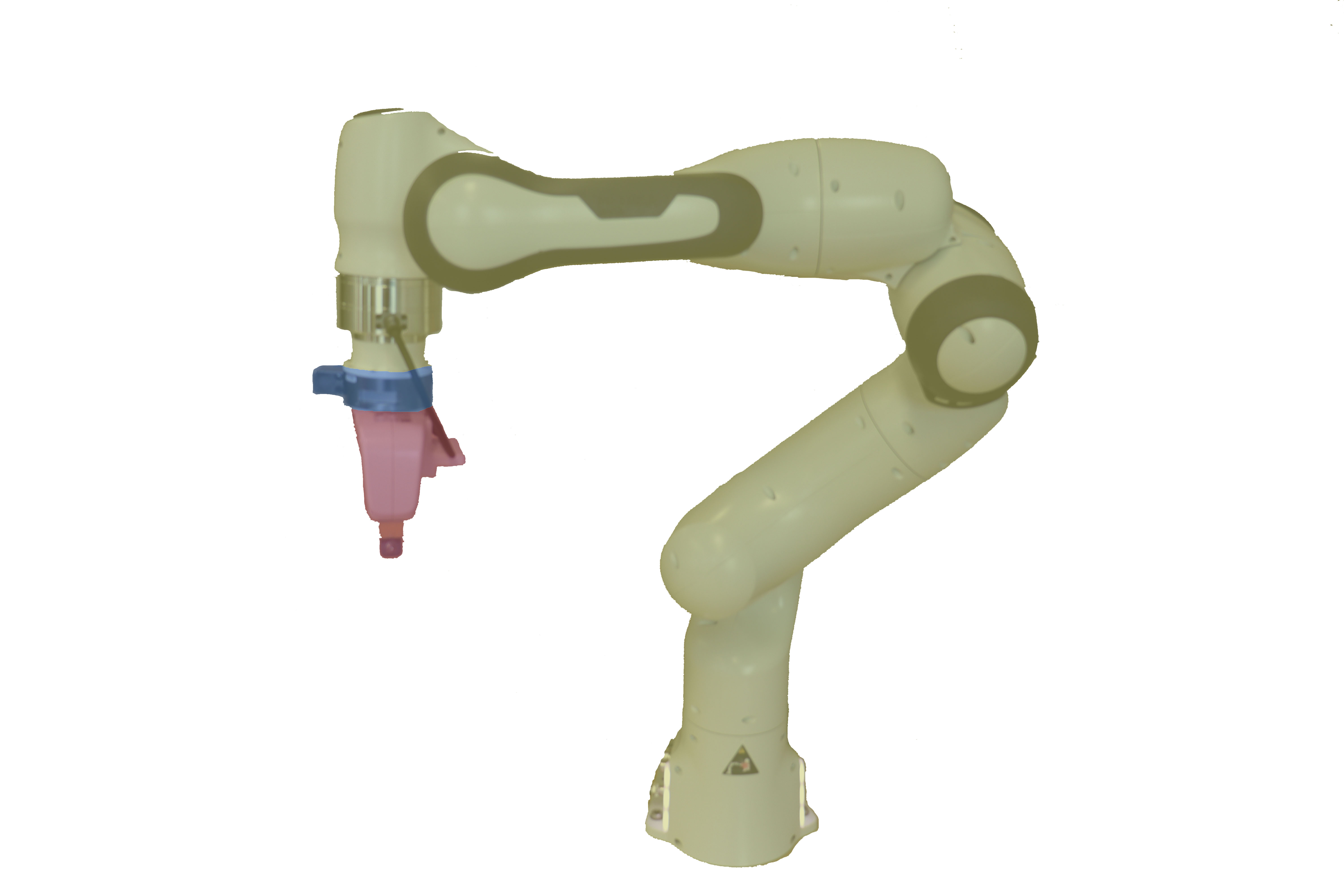}
    \caption{Schematic outline of the setup, showing all measured and calculated parameters (left, \textit{M} denotes the origin of the sensor frame, \textit{P} the payload center of mass), and an exemplary real-world setup from our experiments (right).}
    \label{fig:ple}
\end{figure}

As mentioned in Sect. \ref{sec:introduction}, the measurement of force and torque necessitates the integration of a FTS positioned between the robot and its payload during hand guiding. 
The end-effector pose is utilized to derive the velocities and accelerations of the payload.
The pose data is typically obtained from the robot's control system or calculated via forward kinematics based on the joint configurations. 
Fig. \ref{fig:ple} outlines the described setup. 

First, as stated in \cite{An1988}, estimating payload inertial parameters in Eq. (\ref{eq:newton}) is often approached as a classical optimization problem, typically solved using least-squares (LS) methods. 
These methods assume errors only in dependent variables (forces/torques), ignoring noise in velocities and accelerations, which might necessitate a total least-squares approach (TLS) \cite{Kubus2008}. 
Also, maintaining a measurement rate of at least \SI{100}{\hertz} or using synchronized sensors is stated as critical for accuracy \cite{Farsoni2018}. 
Optimal robot motions, often sinusoidal, minimize jerk, improve data quality, and minimize the condition number of their correlation matrix \cite{Kubus2008}. 
To ensure accurate inertia tensor estimation, these methods require sufficiently strong excitation, especially w.r.t. angular velocities \cite{Traversaro2016}.

A second category of methods utilizes machine learning models, such as neural networks, which approximate the relationship between forces, torques, and inertial parameters \cite{Mavrakis2020}.
These approaches enable parameter estimation from limited interactions (e.g., a single push for a 2D estimation) but depend on large, representative data sets. 
Also data variation due to multiple unknown tools reduces often their reliability. 
Hybrid approaches combine machine learning with classical methods. 
For example, gradient descent on Riemannian manifolds can include a buffer for randomized measurement points, breaking the correlation between consecutive data points \cite{Ledezma2021}. 
Such hybrid approaches ensure physical consistency and the results are comparable to traditional methods without high training data requirements.

Third, image-based methods rely on visual data (e.g., \cite{Buchholz2014,Mirtich1996,Omid2010,Vivek2015,Standley2017}). 
However, these require prior knowledge about the characteristics of the object.
Also, the existing programming system does not provide any camera information. 
Therefore, such methods are less relevant in our context. 

To conclude, while promising, machine learning methods lack analytical approaches' reliability and transparency. 
This, combined with a requirement for large data sets in the initial training phase of such a system, makes them less suitable for our purposes than classical methods. 
For these reasons, we will not explore the application of machine learning to the problem of PIP estimation in this paper any further. 
As the physical consistency of the estimated inertial parameters does not guarantee their correct estimation \cite{Farsoni2021}, it is crucial to ensure sufficient excitation of the payload \cite{Traversaro2016}. 
Therefore, while exploring whether the accelerations during our hand-guided motions are sufficient for a reasonable estimation, we will focus solely on solving the basic optimization problem formulated by Eq. (\ref{eq:newton}) using LS and TLS optimization.

\section{Methodology}
\label{sec:methodology}

As previously stated, our approach for estimating payload inertial parameters is part of an existing non-expert programming framework \cite{Riedl2021,Hartwig2023b,Hartwig2023a} for a Franka Emika Panda robot with a wrist-mounted FTS (Schunk FT model Gamma SI-32-2.5). 
We can derive the direction of the gravity vector and angular and linear velocities from the position and orientation of the FTS. 
Additionally, we can calculate the corresponding accelerations from this. 

For solving the optimization problem (cf. Eq. (\ref{eq:newton})), we employ three different optimization methods.
Reflecting the key issues stated in Sect. \ref{sec:introduction}, we compare them concerning two different measures: (i) the quality of the PIP estimation results they provide (thus measuring how well PIP estimation performs on arbitrary motions) and (ii) their runtime.
First, we use a total least squares (TLS) algorithm similar to the one described by Kubus et al. \cite{Kubus2008} with an additional option to choose between a faster or a more accurate algorithm for singular value decomposition (SVD).
Second, the well-known Ceres solver, which offers automatic differentiation and by default uses a runtime-optimized implementation of the Levenberg-Marquardt algorithm \cite{Agarwal2023}. 
Third, the implementation of the Levenberg-Marquardt algorithm from the Eigen library \cite{Guennebaud2010}, using numerical differentiation. 
For comparison, we perform a brute force sampling of a computable region around the ground truth set of PIP modeled by hand.

\begin{figure}[tb]
    \centering
    \includegraphics[width=0.95\linewidth]{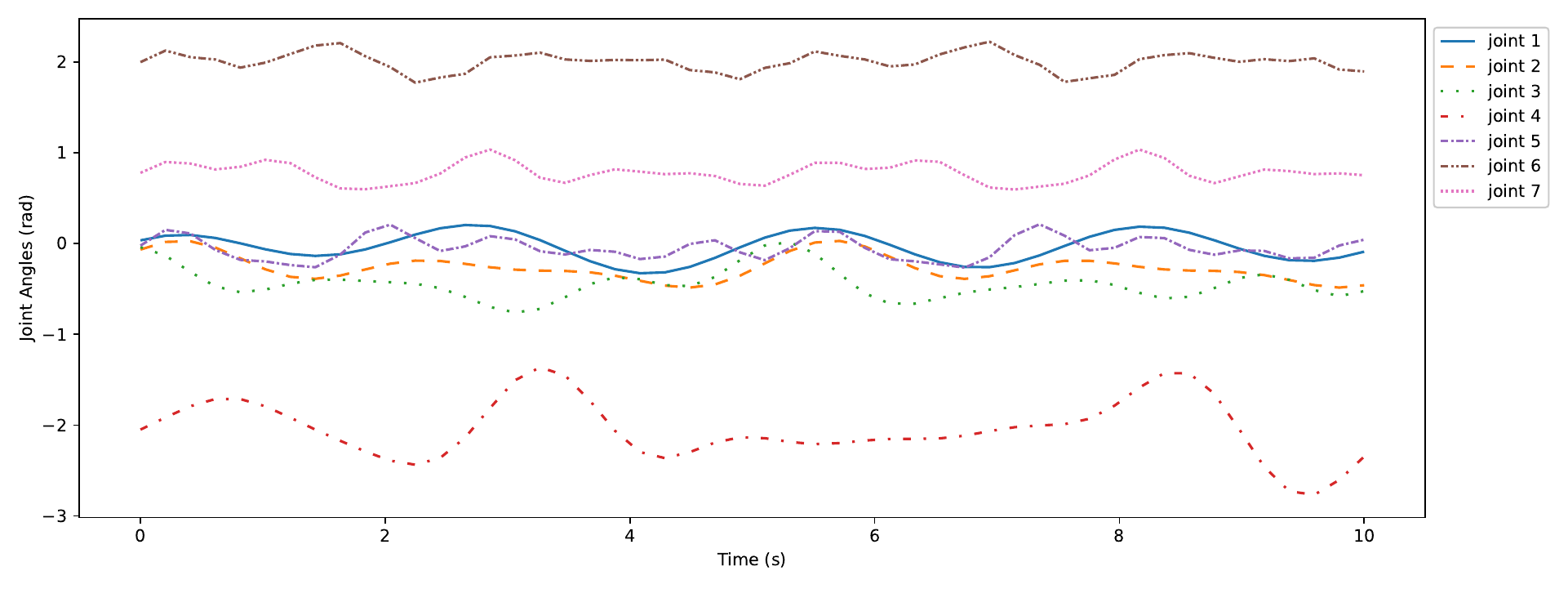}
    \caption{Joint angles for the first \SI{10}{\second} of our predefined motion.}
    \label{fig::motion}
\end{figure}

Our experiments consist of two parts. 
For comparison to established methods as the first part, a predefined motion consisting of a superposition of several weighted sine and cosine functions defining the joint angles (as illustrated in Fig. \ref{fig::motion}) is used to excite a payload mounted directly to the FTS for \SI{20}{\second}. 
This trajectory is optimized to adhere to the limits of the working cell but not to explicitly minimize the condition number of its correlation matrix, as described in \cite{Kubus2008}. 
Payloads used in this part of the experiments are in the range of \SIrange{0.2}{0.5}{\kilogram} of weight, with different centers of mass and inertia tensors.
For the second part, we use the robot's gripper as our payload, amounting to about \SI{0.8}{\kilogram} of weight. 
Instead of the predefined motion, actual hand guiding is performed for \SI{10}{\second}, once in a pick-and-place task and once in a free motion without any task. 
All recorded velocities and forces/torques are sampled at \SI{1}{\kilo\hertz}. 
Both parts investigate the quality of the PIP estimation results. 
Additionally, the first part explores the runtime of each optimization algorithm for our predefined motion.
 
We perform all estimations for mass only, mass and center of mass, and the complete set of PIP, using our modeled ground truth data for any parameters that are not estimated. 
Due to discretization by the robot's controller, the measured velocities are filtered before PIP estimation (using Savitzky–Golay filtering \cite{Savitzky1964} with a third-order polynomial and a window width of ten). However, we also briefly explore stronger filtering on the complete data set. 
In order to be usable with any robot trajectory, our PIP estimation methods also generally discard the first and last 10 percent of the data, making smooth introductory and ending phases of the hand-guided motions unnecessary.

The evaluation of both parts consists of a validation step and an actual PIP estimation step. 
Using the ground truth parameters modeled by hand, forces and torques are calculated according to Eq. (\ref{eq:newton}) from the measured velocities and corresponding accelerations. 
We then feed this validation data into our different optimization algorithms.
PIP estimation on these data sets shows the theoretically possible quality of the results provided by our chosen estimation methods as well as their runtime. 
In the actual PIP estimation step, the real measured data is used for forces and torques, allowing us to evaluate how sensitive the different estimation methods are to noise.

To summarize, our evaluation focuses on the quality of the results obtained from each PIP estimation and the runtime of the different optimization methods we employ. 
For all error measures, we use a relative error $e$ between the estimated parameter $p_{\text{e}}$ and the modeled ground truth parameter $p_{\text{gt}}$ defined as:
\begin{equation}
    e = \frac{|| (p_{\text{e}} - p_{\text{gt}})||_2}{||p_{\text{gt}}||_2}
\label{eq:error}
\end{equation}
where $p$ is either the mass $m$, center of mass $c$, or inertia tensor $I$, and $||p||_2$ denotes the corresponding L2 norm of a given parameter $p$.

\section{Results}
\label{sec:results}

\begin{figure}[tb]
    \centering
    \begin{subfigure}[b]{0.49\linewidth}
        \includegraphics[width=\linewidth]{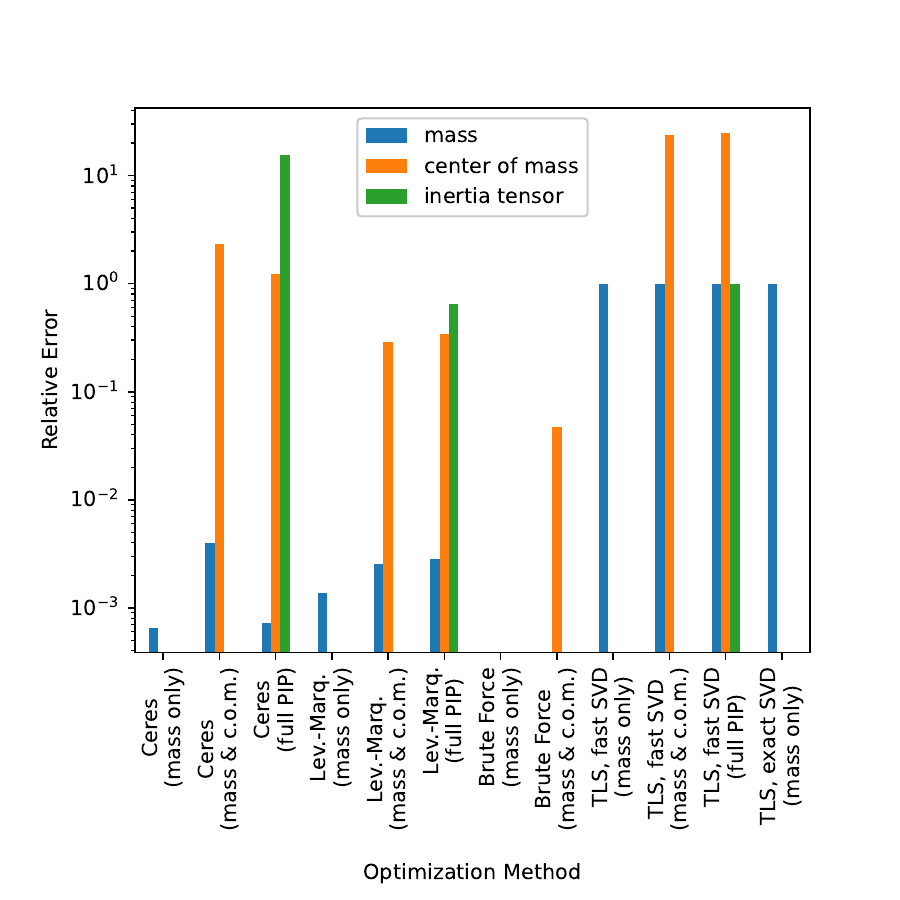}
        \caption{}
    \end{subfigure}
    \begin{subfigure}[b]{0.49\linewidth}
        \includegraphics[width=\linewidth]{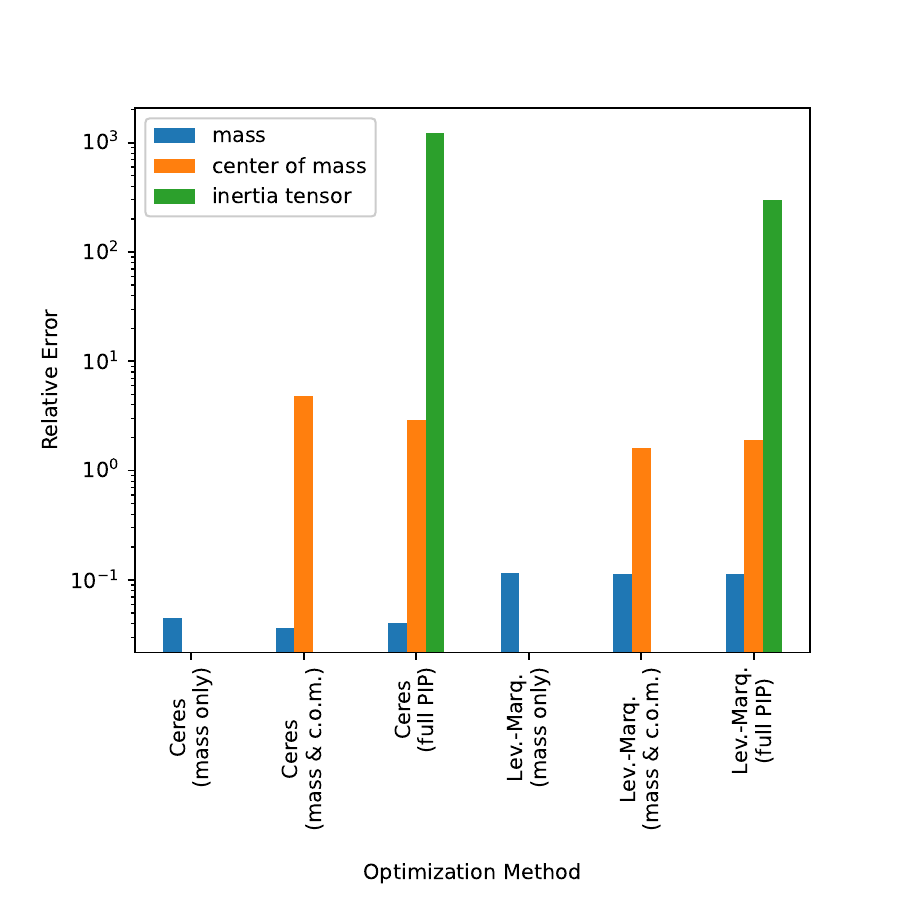}
        \caption{}
    \end{subfigure}
    \caption{Relative error for the different optimization methods, using (a) validation data calculated from velocity measurements, and (b) actual measured data from our predefined motion (note the different scales necessary for visualization).}
    \label{fig::synth}
\end{figure}

When looking at the PIP estimation results for our predefined motion (see Fig. \ref{fig::synth}), the largely different order of magnitude in errors for the validation data set and the actual measured data (a factor of up to and for inertia tensor $I$ even greater than $10^2$, cf. Table \ref{tab::err}) is apparent immediately. 
This finding indicates how much influence the input data quality has on the quality of the estimation results. 
Using our validation data, the TLS approach performs much worse than ordinary least-squares methods, starkly contrasting results like those reported by Kubus et al. \cite{Kubus2008}.
This is true not only in terms of PIP estimation quality but also when it comes to algorithm runtime. 
Even using only every tenth sample from the input data, our TLS implementation takes too long to perform estimations that the other algorithms finish within seconds while using the entire data sets (cf. Table \ref{tab::runtime}).
For this reason, we decided not to evaluate our TLS algorithm beyond the estimation results on the validation data from the first part of the experiment.
\begin{table}[tb]
    \centering
    \caption{Relative error values for least-squares PIP estimation performed on the data gathered from our predefined motion (both validation and measured data)}
    \label{tab::err}
    \begin{tabular}{|P{1.5cm}|P{1.5cm}|P{1.5cm}|P{1.5cm}|P{1.5cm}|P{1.5cm}|P{1.5cm}|}
        \hline
         & Ceres (mass only) & Ceres (mass \& c.o.m.) & Ceres (full PIP) & Eigen Lev.-Marq. (mass only) & Eigen Lev.-Marq. (mass \& c.o.m.) & Eigen Lev.-Marq. (full PIP) \\
         \hline
         $m_{\text{val}}$ & 0.00065 & 0.00397 & 0.00072 & 0.00139 & 0.00253 & 0.00283 \\
         \hline
         $m_{\text{meas}}$ & 0.04469 & 0.03657 & 0.04058 & 0.11658 & 0.11369 & 0.11303 \\
         \hline
         $c_{\text{val}}$ & - & 2.32594 & 1.22541 & - & 0.29014 & 0.34524 \\
         \hline
         $c_{\text{meas}}$ & - & 4.78282 & 2.89111 & - & 1.61306 & 1.89158 \\
         \hline
         $I_{\text{val}}$ & - & - & 15.4697 & - & - & 0.65533 \\
         \hline
         $I_{\text{meas}}$ & - & - & 1232.35 & - & - & 297.511 \\
         \hline
    \end{tabular}
\end{table}

\begin{table}[tb]
    \centering    
    \caption{Runtime of the different PIP estimation methods (no entry means PIP estimation was aborted due to excessive runtime)}
    \label{tab::runtime}
    \begin{tabular}{|P{3cm}|P{2cm}|P{2cm}|P{2cm}|}
        \hline
        Optimization method & mass only & mass and c.o.m. & full PIP \\
        \hline
         Ceres & \textless \SI{1}{\second} & \textless \SI{1}{\second} & \SI{1}{\second} \\ 
         \hline
         Eigen Lev.-Marq. & \SI{1}{\second} & \SI{4}{\second} & \SI{45}{\second} \\ 
         \hline
         TLS (fast SVD) & \SI{2.9}{\hour} & \SI{41.1}{\hour} & \SI{71}{\hour} \\ 
         \hline
         TLS (exact SVD) & \SI{20}{\hour} & - & - \\ 
         \hline
         Brute Force & \textless \SI{1}{\second} & \SI{1.5}{\minute} & - \\ 
         \hline
    \end{tabular}
\end{table}
Another point that becomes evident immediately is that for all optimization methods, the proper estimation of the inertia tensor $I$ proves to be very prone to errors, being the most error-prone of the PIP by far, mainly when calculating estimations from actual measured data.
As the quality of the measured data is decisive for the quality of PIP estimation results, we also used the data from our predefined motion to test the effect of different smoothing methods. 
However, this kind of filtering led to worse estimation results for our data, especially for the center of mass $c$ and the inertia tensor $I$.
This indicates that the optimization algorithms we implemented are already capable of reducing the influence of structured noise and that stronger filtering of the input is not helpful for the quality of PIP estimation.

\begin{figure}[tb]
    \centering
    \begin{subfigure}[b]{0.49\linewidth}
        \includegraphics[width=\linewidth]{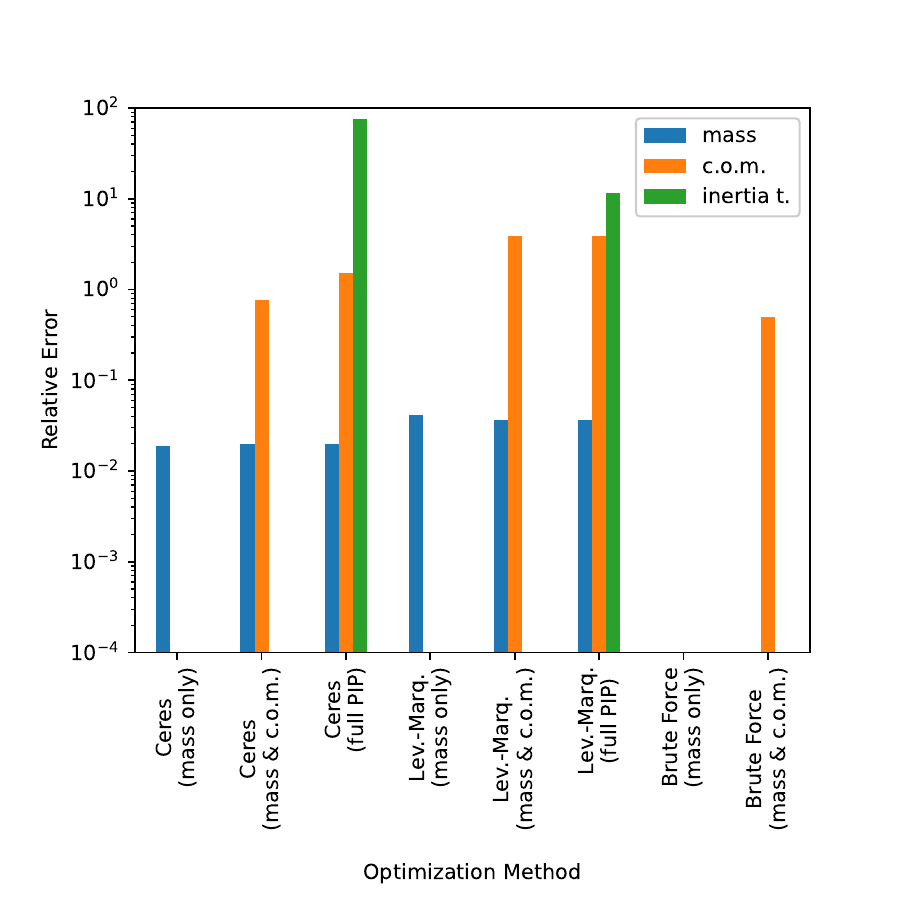}
        \caption{}
    \end{subfigure}
    \begin{subfigure}[b]{0.49\linewidth}
        \includegraphics[width=\linewidth]{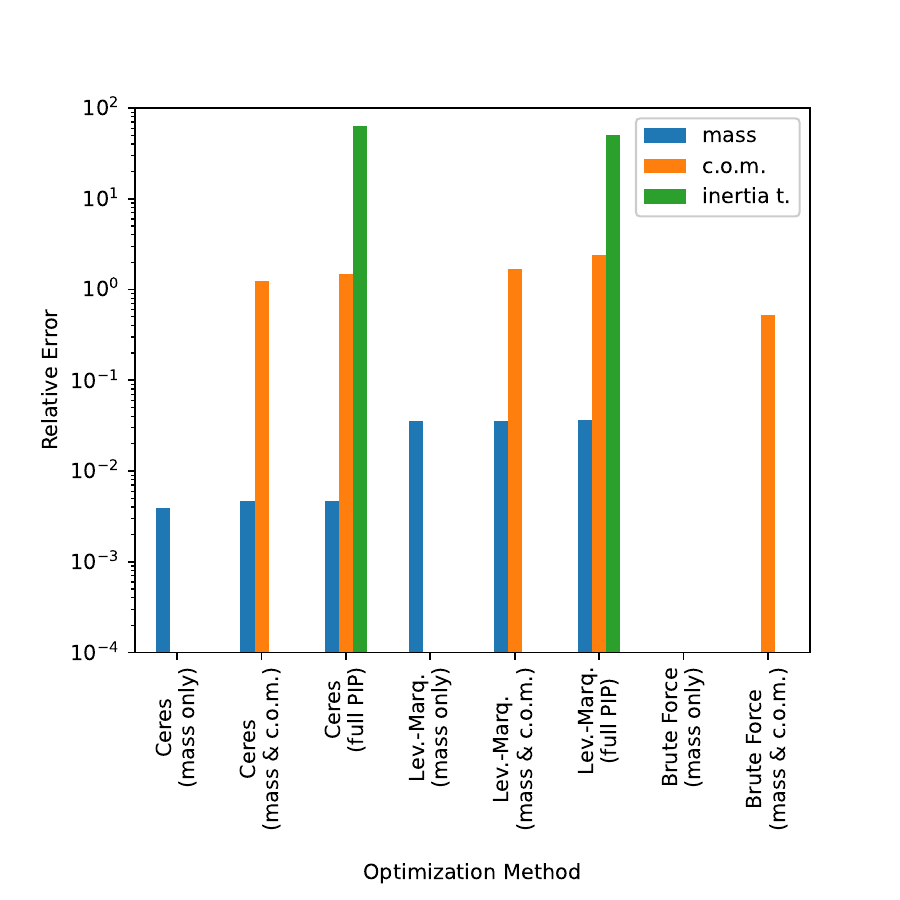}
        \caption{}
    \end{subfigure}
    \caption{Relative error for the different optimization methods, using actual measured data for (a) the simulated pick-and-place task and (b) hand guiding without a predefined task or target.}
    \label{fig::kin_rel}
\end{figure}

Evaluating our data gathered from hand guiding, we see a similar, albeit smaller, difference in the errors for validation and actual measured data. 
Errors for both the pick-and-place task and the free motion are relatively similar, especially for actual measured data (cf. Fig. \ref{fig::kin_rel}). 
Once again, the mass $m$ is estimated to have the highest accuracy, while the inertia tensor $I$ notably differs from the model data.

\begin{figure}
    \centering
    \begin{subfigure}[b]{0.49\linewidth}
        \includegraphics[width=0.98\linewidth]{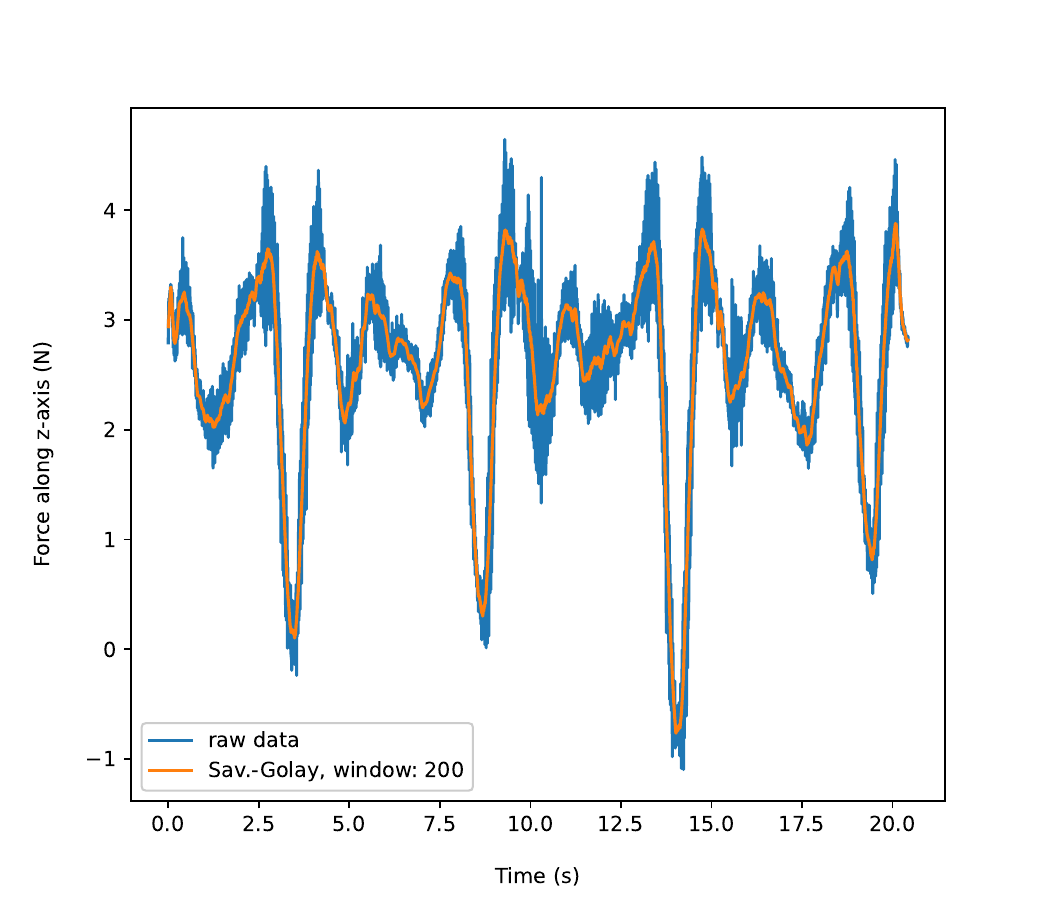}
        \caption{}
    \end{subfigure}
    \begin{subfigure}[b]{0.49\linewidth}
        \includegraphics[width=0.98\linewidth]{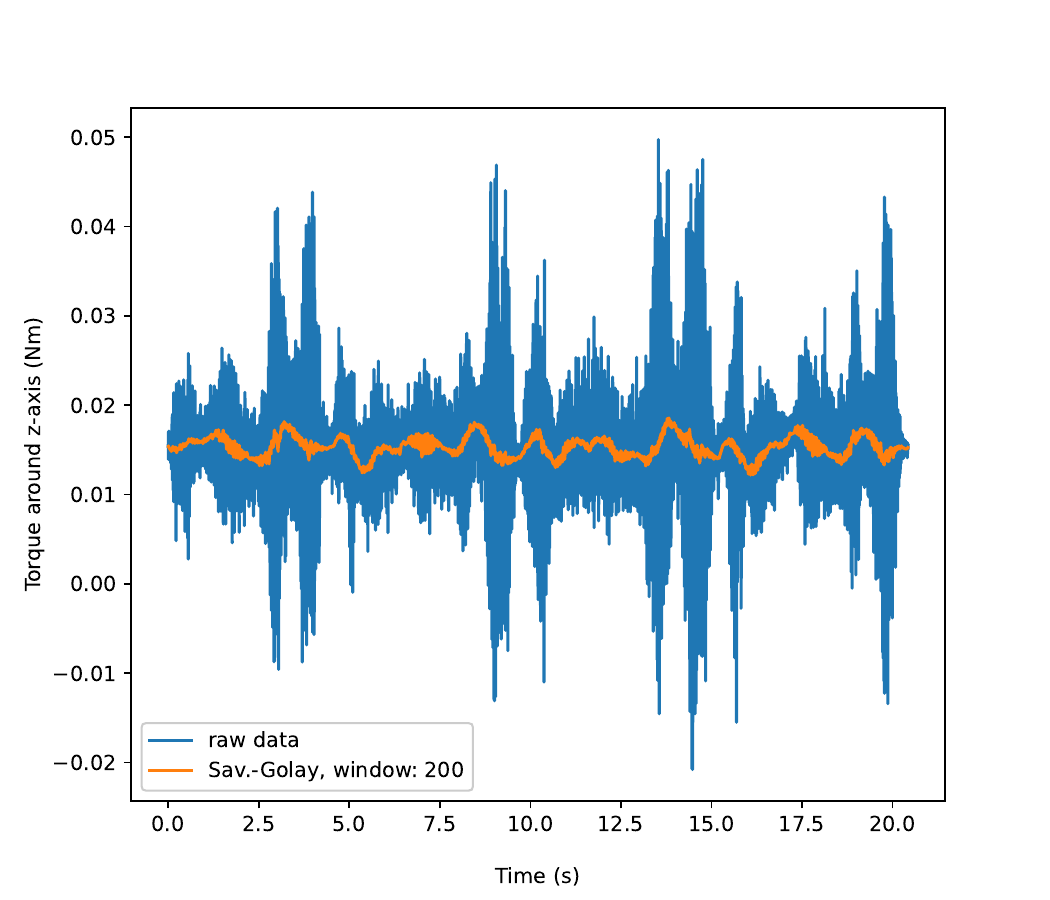}
        \caption{}
    \end{subfigure}
    \caption{Measurements for (a) force along and (b) torque around the z-axis during execution of our predefined motion.}
    \label{fig:data}
\end{figure}

Overall, the higher errors when estimating the inertia tensor $I$ and, to some extent, also the center of mass $c$ showcase a key issue that comes with using arbitrary motions in PIP estimation. 
Specifically designed trajectories emphasize on ensuring sufficient excitation of the payload \cite{Kubus2008}, a criterion we cannot enforce in our use case. 
As shown in Fig. \ref{fig:data}, this leads to significant differences between the measurements for different degrees of freedom, most notably between forces and torques, especially along the z-axis. 
Since $I$ is estimated exclusively from the measured torques and corresponding angular velocities and accelerations, our data does not provide sufficient information for an exact PIP estimation. 
Without enough angular acceleration, the elements of the inertia tensor $I$ can even become not identifiable numerically \cite{Traversaro2016}. 
The mass $m$, on the other hand, can easily be estimated from the measured forces and corresponding linear acceleration (including acceleration by gravity), and for estimating the center of mass $c$, information of both forces and torques is combined, still leading to usable but inaccurate results. 
This shows that not all hand guiding motions are equally applicable, and a selection should be made beforehand.

\section{Conclusion}
\label{sec:conclusion}

The results show that estimating payload inertial parameters from motions demonstrated by hand guiding is both (i) generally feasible and (ii) possible within a reasonable time frame, addressing our key research issues. 
Accurately estimating the inertia tensor $I$ remains a significant challenge due to the low angular velocities, as seen in our data. 
Consequently, PIP estimation from arbitrary motions, as performed in this study, cannot match the high quality of the results of methods that employ specifically designed trajectories.

However, it should be noted that our concept exclusively uses sensor data, assuming that the human interaction in our programming scenario is limited only to the hand guiding of the robot. 
With additional input from the user, e.g., picking a shape that best matches the payload from a list of available options, the inertia tensor $I$ could be estimated more accurately, reducing the overall error. 
Furthermore, users who are domain experts can provide even more detailed information on the payload and its physical characteristics, shifting the focus from full PIP estimation to validation or iterative on-line updates of their inputs. 
Nevertheless, this might be slowing down the programming process.

Additionally, future research could explore the application of machine learning techniques, particularly data-driven approaches. 
Such methods might address the lack of information inherent in the torque, angular velocity, and acceleration data when utilizing motions from hand guiding. 
Another research direction could be adapting techniques like those by Farsoni et al. \cite{Farsoni2021}. 
We could add an excitation trajectory to a demonstrated non-contact motion, ensuring the motion follows safety guidelines for human-robot interaction during reproduction.

%
%

\end{document}